\documentclass[11pt,letterpaper]{article}

\usepackage{naaclhlt2015}
\usepackage{times}
\usepackage{url}
\usepackage{latexsym}
\usepackage[utf8]{inputenc} 
\usepackage[T1]{fontenc}
\usepackage{graphicx}
\usepackage{amsmath}
\usepackage{comment}
\usepackage[small]{caption}

\usepackage{amssymb}
\usepackage{algorithm}
\usepackage[noend]{algpseudocode}
\usepackage{titlesec}
\usepackage{dblfloatfix} 
\usepackage{caption}
\usepackage{url}
\usepackage{tipa}
\usepackage[table]{xcolor}
\usepackage{xspace}

\long\def\eat#1{\ignorespaces}
\newcommand{\enotesoff}{\long\gdef\enote##1##2{}}

\enotesoff

\newcommand*\NoIndentAfterEnv[1]{%
  \AfterEndEnvironment{#1}{\par\@afterindentfalse\@afterheading}}
\makeatother

\newif\ifverbose
\verbosefalse

\ifverbose
\usepackage{todonotes} 
\newcommand{\topic}[1]{\paragraph{\colorbox{pink}{{\small {\sc topic}: #1}}}\hfill\newline}
\newcommand{\rationale}[1]{\noindent{\footnotesize \colorbox{green}{{\sc rationale}:} #1}\hfill\newline}

\else
\usepackage[disable]{todonotes}
\newcommand{\topic}[1]{}
\newcommand{\rationale}[1]{}
\fi

\newcommand{\morphosim}{\textsc{MorphoSim}\xspace}

\newcommand{\word}[1]{\texttt{#1}}
\newcommand{\tagf}[1]{\textsc{#1}}
\newcommand{\noun}{\textsc{Noun}}

\newcommand{\defeq}{\stackrel{\mbox{\tiny def}}{=}}

\title{\raisebox{1ex}[0in][0in]{\parbox[b]{\linewidth}{\begin{flushright}\footnotesize
        \textmd{\textsf{\textcolor{gray}{Appeared in the
                proceedings of NAACL 2015 (Denver, June).  This
          version was \\ prepared in June 2019 and fixes a few typos.}}}\end{flushright}}}\\ \vspace{-1ex}Morphological Word Embeddings}

\author{Ryan Cotterell$^{1,2}$ \\
  Department of Computer Science$^{1}$ \\
  Johns Hopkins University, USA \\
  {\tt ryan.cotterell@jhu.edu}
  \And
  Hinrich Sch{\"u}tze$^{2}$ \\
  Center for Information and Language Processing$^2$ \\
  University of Munich, Germany  \\
  {\tt inquiries@cislmu.org}
}

\begin{document}
\maketitle

\begin{abstract}
Linguistic similarity is multi-faceted. For instance, two words may be
similar with respect to semantics, syntax, or morphology {\em inter alia}.
Continuous word-embeddings have been shown to capture most of these
shades of similarity to some degree. This work considers guiding
word-embeddings with morphologically annotated data, a form of
semi-supervised learning, encouraging the vectors to encode a word's
morphology, i.e., words close in the embedded space share
morphological features. We extend the log-bilinear model
to this end
and show that indeed our learned embeddings achieve this, using
German as a case study.
\end{abstract}

\section{Introduction}\label{sec:introduction}

\topic{broad-brush-stroke attack on bag of words from a morphological
angle}

Word representation is fundamental for NLP. Recently, continuous
word-embeddings have gained traction as a general-purpose
representation framework. While such embeddings have proven themselves
useful, they typically treat words holistically, ignoring their
internal structure. For morphologically impoverished languages, i.e., languages with a
low morpheme-per-word ratio such as English, this is often not a
problem.  However, for the processing of morphologically-rich
languages exploiting word-internal structure is necessary.

\topic{bird's eye view of our approach}

Word-embeddings are typically trained to produce representations that
capture linguistic similarity. The general idea is that words that are
close in the embedding space should be close in meaning.  A key
issue, however, is that meaning is a multi-faceted concept and thus
there are multiple axes, along which two words can be similar. For example,
\word{ice} and \word{cold} are \emph{topically} related, \word{ice} and
\word{fire} are \emph{syntactically} related as they are both nouns, and
\word{ice} and \word{icy} are \emph{morphologically} related as they are
both derived from the same root. In this work, we are interested in
distinguishing between these various axes and guiding the embeddings
such that similar embeddings are morphologically related.
\topic{informal overview of the formal details}

We augment the 
log-bilinear model (LBL) of \newcite{mnih2007three}
with a multi-task objective. In addition to raw text, our model is trained
on a corpus annotated with morphological tags, encouraging
the vectors to encode a word's morphology. To be concrete,
the first task is language modeling---the traditional use of
the LBL---and the second is akin to unigram morphological tagging. The LBL,
described in section \ref{sec:log-bilinear-model}, is fundamentally a
language model (LM)---word-embeddings fall out as low dimensional
representations of context used to predict the next word. We extend
the model to jointly predict the next morphological tag along with the
next word, encouraging the resulting embeddings to encode morphology.  We present a novel
metric and experiments on German as a case study that demonstrates that our
approach produces word-embeddings that better preserve morphological
relationships.

\enote{hs}{i changed the caption because the phrase is not
  from TIGER, I guess?}

\eat{
\begin{table*}
\scalebox{.95}{
\begin{tabular}{|c|c|c|c|c|} \hline
\rowcolor{blue!10} \tagf{Article} & \tagf{Adjective} & \tagf{Noun} & \tagf{Article} & \tagf{Noun} \\ \hline
\rowcolor{blue!20} {\small \tagf{Art.Def.Nom.Sg.Fem}} &  {\small \tagf{Adj.Nom.Sg.Fem}} &{\small \tagf{N.Nom.Sg.Fem}} & {\small \tagf{Art.Def.Gen.Sg.Fem}} & {\small \tagf{N.Gen.Sg.Fem} }\\ \hline
\rowcolor{green!10} \word{die} & \word{gr{\"o}{\ss}te} & \word{Volkswirtschaft} & \word{der} & \word{Welt} \\ \hline
\rowcolor{green!20} the & biggest & economy & in the & world \\ \hline
\end{tabular}}
\caption{A sample German sentence from the TIGER corpus with an
accompanying English translation. Each word is annotated with a complex
morphological tag and its corresponding coarse-grained POS tag. For
instance, \word{Welt} is annotated with \tagf{N.Gen.Sg.Fem} indicating
that it is a {\em noun} in the {\em genitive} case and also both
{\em singular} and {\em feminine}. Note that each tag is composed of
meaningful sub-tag units that are shared across whole tags,
e.g, the nominative feature fires on both adjectival and nominal
tags.}
\label{table:german-example}
\end{table*}
}

\topic{Review of morphology in language models}

\enote{hs}{i put a second subsection heading in to make full
  use of the four pages -- remove if you run out of sapce}

\section{Related Work}\label{sec:related-work}
Here we discuss the role morphology has played
in language modeling and offer a brief overview
of various approaches to the larger task of computational morphology. 

\subsection{Morphology in Language Modeling}
Morphological structure has been previously integrated into LMs. Most notably, \newcite{bilmes2003factored} introduced {\em
factored LMs}, which effectively add tiers, allowing easy
incorporation of morphological structure as well as
part-of-speech (POS) tags.  More
recently, \newcite{mueller2011improved} trained a class-based LM using common suffixes---often indicative of
morphology---achieving state-of-the-art results when interpolated with
a Kneser-Ney LM.  In neural probabilistic modeling,
\newcite{luong2013better} described a recursive neural network
LM, whose topology was derived from the output of
\textsc{Morfessor}, an unsupervised morphological segmentation tool
\cite{creutz2005unsupervised}. Similarly, \newcite{qiu2014}
augmented \textsc{word2vec} \cite{mikolov2013efficient} to embed morphs as well as whole
words---also taking advantage of \textsc{Morfessor}.
LMs were tackled by
\newcite{dos2014learning}
with a
convolutional neural network with a $k$-best max-pooling layer to
extract character level $n$-grams, efficiently inserting orthographic
features into the LM---use of the vectors in down-stream POS
tagging achieved state-of-the-art results in Portuguese. Finally, most
similar to our model, \newcite{Botha2014} introduced the additive
log-bilinear model (LBL++). Best summarized as a neural factored
LM, the LBL++ created separate embeddings for each
constituent morpheme of a word, summing them to get a single
word-embedding.

\begin{table}
\setlength\tabcolsep{.15cm}
{\small
\begin{tabular}{|c|c|c|c|c|} \hline
\rowcolor{blue!10} \tagf{Article} & \tagf{Adjective} & \tagf{Noun}  \\ \hline
\rowcolor{blue!20} {\scriptsize \tagf{Art.Def.Nom.Sg.Fem}} &  {\scriptsize \tagf{Adj.Nom.Sg.Fem}} &{\scriptsize \tagf{N.Nom.Sg.Fem}} \\ \hline
\rowcolor{green!10} \word{die} & \word{gr{\"o}{\ss}te} & \word{Stadt} \\ \hline
\rowcolor{green!20} the & biggest & city \\ \hline
\end{tabular}
}
\caption{A sample German phrase in TIGER \cite{tiger} annotation with an
accompanying English translation. Each word is annotated with a complex
morphological tag and its corresponding coarse-grained POS tag. For
instance, \word{Stadt} is annotated with \tagf{N.Nom.Sg.Fem} indicating
that it is a {\em noun} in the {\em nominative} case and also both
{\em singular} and {\em feminine}. Each tag is composed of
meaningful \emph{sub-tag units} that are shared across whole tags,
e.g., the feature \tagf{Nom} fires on both adjectives and nouns.}
\label{table:german-example}
\end{table}

\subsection{Computational  Morphology}\label{sec:morph-tagging}

\topic{Introduce morphological tagging.}

Our work is also related to morphological tagging, which can be
thought of as ultra-fine-grained POS tagging. For
morphologically impoverished languages, such as English, it is natural to consider a
small tag set. For instance, in their universal POS tagset,
\newcite{petrov2011universal} propose the coarse tag \noun\,to
represent all substantives. In inflectionally-rich languages, like
German, considering other nominal attributes, e.g., case, gender and
number, is also important. An example of an annotated German phrase is found in
table \ref{table:german-example}. This often leads to a
large tag
set; e.g., in the morphological tag set of
\newcite{hajic2000morphological}, English had 137 tags whereas
morphologically-rich Czech had 970 tags!

\topic{discuss distributional morphology}
Clearly, much of the information needed to determine a word's
morphological tag is encoded in the word itself. For example, the suffix
\word{ed} is generally indicative of the past tense in
English. However, distributional similarity has also been shown to be
an important cue for morphology
\cite{yarowsky2000minimally,schone2001knowledge}. Much as contextual
signatures are reliably exploited approximations to the semantics of
the lexicon \cite{harris1954distributional}---{\em you shall know the
meaning of the word by the company it keeps} \cite{firth57}---they
can be similarly exploited for morphological analysis. This is not an
unexpected result---in German, e.g., we would expect nouns that
follow an adjective in the genitive case to also be in the genitive
case themselves.  Much of what our model is designed to accomplish is
the isolation of the components of the contextual signature that are
indeed predictive of morphology.

\section{Log-Bilinear Model}\label{sec:log-bilinear-model}

The LBL is a generalization of the
well-known log-linear model. The key difference lies in how
it deals with
features---instead of making use of hand-crafted features, the
LBL {\em learns} the features along with the weights. In the
language modeling setting, we define the following model,
\begin{equation} p(w\mid h) \defeq
\frac{\exp\left(s_\theta(w,h)\right)}{\sum_{w'}
\exp\left(s_\theta(w',h)\right)},
\end{equation} where $w$ is a word, $h$ is a history and $s_\theta$ is
an energy function. Following the notation
of \newcite{mnih2012}, in the LBL we define
\begin{equation} s_\theta(w,h) \defeq \left(\sum_{i=1}^{n-1} C_i
r_{h_i}\right)^T q_w + b_w,
\end{equation} where 
$n-1$ is the history length 
and
the parameters $\theta$ consist of $C$, a matrix
of context specific weights, $R$, the context word-embeddings, $Q$,
the target word-embeddings, and $b$, a bias term. Note that a
subscripted matrix indicates a vector, e.g., $q_w$ indicates the
target word-embedding for word $w$ and $r_{h_i}$ is the embedding for
the $i$th word in the history. The gradient, as in all energy-based
models, takes the form of the difference between two expectations
\cite{lecun2006tutorial}.

\topic{Overview of the model}
\section{Morph-LBL}

We propose a multi-task objective that jointly
predicts the next word $w$ and its morphological tag $t$ given a
history $h$. Thus we are interested in a joint probability distribution
defined as
\begin{equation}
  p(w, t \mid h) \propto \exp(( f_t^T S  + \sum_{i=1}^{n-1}C_i r_{h_i})^T q_w + b_{w} )
\end{equation}
where $f_t$ is a hand-crafted feature vector for a morphological tag
$t$  and $S$ is an additional weight
matrix. Upon inspection, we see that
\begin{equation}
p(t \mid w,h) \propto \exp(S^T f_t q_w)
\end{equation}
Hence given a fixed embedding $q_w$ for word $w$, we can interpret
$S$ as the weights of a conditional log-linear model used to predict
the tag $t$.

\topic{Describe feature function.}
Morphological tags lend themselves to easy featurization. As shown in
table \ref{table:german-example}, the morphological tag
\textsc{Adj.Nom.Sg.Fem} decomposes into sub-tag units \textsc{Adj},
\textsc{Nom}, \textsc{Sg} and \textsc{Fem}. Our model
includes a binary feature for each sub-tag unit in the tag set 
and only those present in a given tag fire; e.g., $F_{\textsc{Adj.Nom.Sg.Fem}}$
is a vector with exactly four non-zero components.

\subsection{Semi-Supervised Learning}

In the fully supervised case, the method we proposed above requires a corpus
annotated with morphological tags to train. This conflicts
with a key use case of word-embeddings---they allow the easy incorporation
of large, unannotated corpora into supervised tasks \cite{turian2010word}.
To resolve this, we train our model on a partially annotated corpus. 
The key idea here is that we only need a partial set of labeled data
to steer the embeddings to ensure they capture morphological properties of 
the words. We marginalize out the tags for the subset of the
data
\enote{hs}{, no comma for restrictive relative clause?}
for which we do not have annotation. 

\section{Evaluation}

\begin{figure}
\includegraphics[width=.5\textwidth]{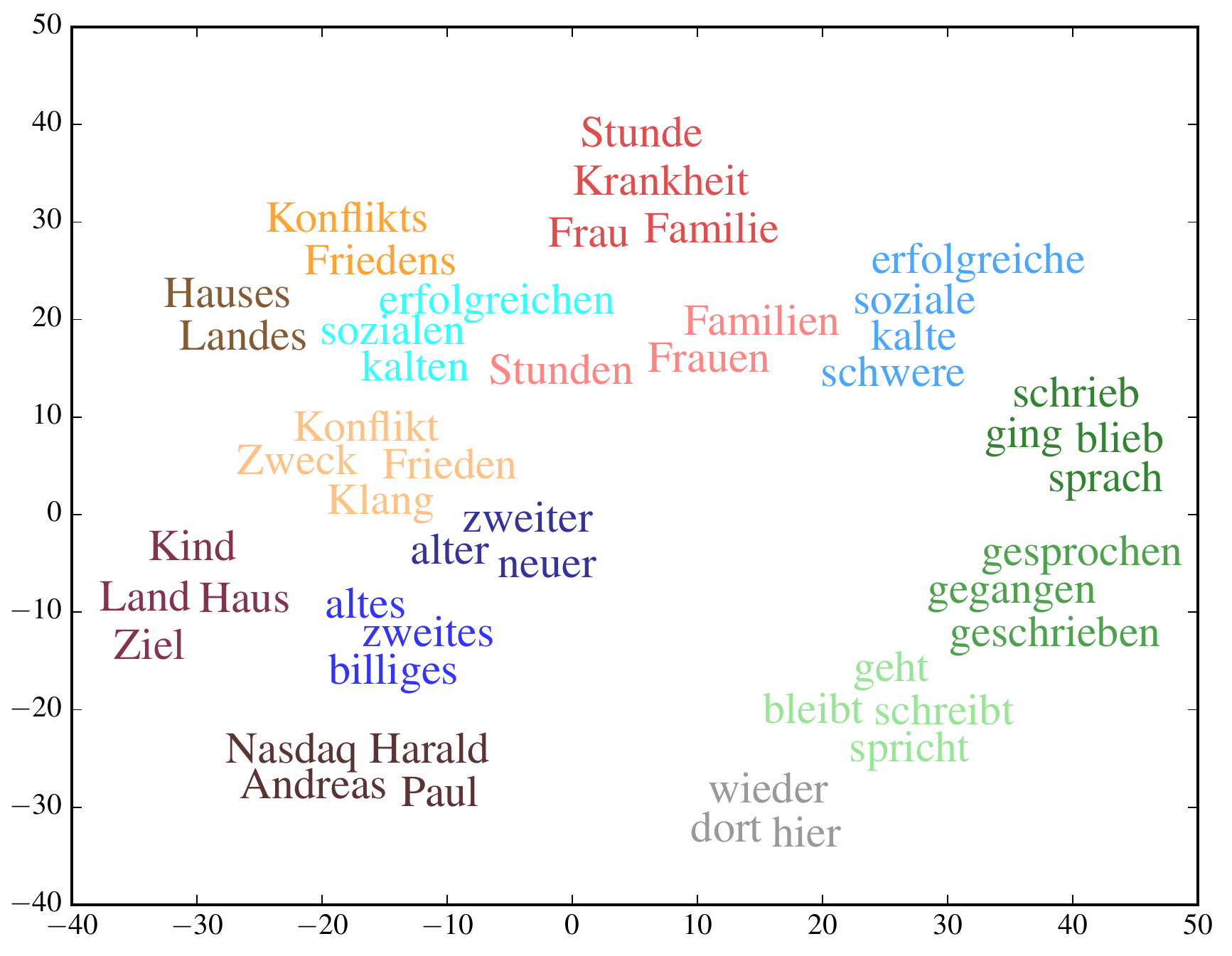}
\caption{Projections of our 100 dimensional embeddings onto
$\mathbb{R}^2$ through t-SNE \cite{van2008visualizing}. Each word is
given a distinct color determined by its morphological tag. 
We see clear clusters reflecting morphological tags and coarse-grained POS---verbs are in
various shades of green, adjectives in blue, adverbs in grey and nouns
in red and orange. Moreover, we see similarity across coarse-grained
POS tags, e.g., the genitive adjective \word{sozialen} lives near the
genitive noun \word{Friedens}, reflecting the fact that
``sozialen Friedens'' `social peace' is a frequently used
German phrase.\label{fig:tsne}}
\end{figure}

In our evaluation, we
attempt to intrinsically determine whether 
it is indeed true that words similar in the embedding
space are morphologically related. 
Qualitative evaluation, shown
in figure \ref{fig:tsne}, indicates that
this is the case.

\subsection{MorphoSim}\label{sec:morpho-sim}

We introduce a new evaluation metric for morphologically-driven
embeddings to quantitatively score models. Roughly, the question
we want to evaluate is: are words that are similar in the embedded
space also morphologically related? Given a word $w$ and its
embedding $q_w$, let $\mathcal{M}_w$ be the set of morphological tags
associated with $w$ represented by bit vectors. This is a
set because words may have several morphological parses.
Our measure is then defined below,
\begin{equation*} \morphosim(w) \defeq 
-\sum_{w' \in \mathcal{K}_w}
\min_{m_{w}, m_{w'}} d_h(m_w,m_{w'})
\end{equation*} where $m_{w} \in \mathcal{M}_{w}$, $m_{w'} \in
\mathcal{M}_{w'}$, $d_h$ is the Hamming distance and $\mathcal{K}_w$ is
a set of words close to $w$ in the embedding space. We are given some
freedom in choosing the set $\mathcal{K}_w$---in our experiments we take
$\mathcal{K}_w$ to be the $k$-nearest neighbors ($k$-NN) in the embedded space using
cosine distance. We report performance under this evaluation metric
for various $k$. Note that \morphosim can be viewed as a soft
version of $k$-NN---we measure not just whether a word has the same
morphological tag as its neighbors, but rather has a {\em similar} morphological tag.

Metrics similar to \morphosim have been
applied in the speech recognition community. For example, \newcite{levin2013fixed}
had a similar motivation for their evaluation of fixed-length acoustic embeddings
that preserve linguistic similarity.

\section{Experiments and Results}
To show the potential of our approach, we chose to perform a case
study on German, a morphologically-rich language.  We conducted
experiments on the TIGER corpus of newspaper German
\cite{tiger}. To the best of our knowledge, no previous
word-embedding techniques have attempted to incorporate {\em morphological
tags} into embeddings in a supervised fashion. We note again that there
has been recent work on incorporating {\em morphological segmentations}
into embeddings---generally in a pipelined approach using a segmenter,
e.g., \textsc{Morfessor}, as a preprocessing step, but we distinguish our
model through its use of a different view on morphology. 

We opted to
compare Morph-LBL with two fully unsupervised models: the original LBL
and \textsc{word2vec} 
({\small
\url{code.google.com/p/word2vec/}}, \newcite{mikolov2013efficient}). All models were trained on the first 200k
words of the train split of the TIGER corpus; Morph-LBL was given the
correct morphological annotation for the first 100k words.  The LBL and
Morph-LBL models were implemented in Python using \textsc{theano}
\cite{Bastien-Theano-2012}. All vectors had dimensionality 200. We
used the Skip-Gram model of the \textsc{word2vec} toolkit with context
$n=5$.
We initialized
parameters of
LBL and Morph-LBL randomly and trained them using
stochastic gradient descent \cite{robbins1951stochastic}. 
We used a history size of $n=4$.
\subsection{Experiment 1: Morphological Content}
\begin{table}
\center
\begin{tabular}{|l|||c|c|c|c|} \hline
& Morph-LBL & LBL & \textsc{word2vec} \\ \hline
All Types & 81.5\% & 22.1\% & 10.2\% \\ \hline
No Tags & 44.8\% & 15.3\%& 14.8\% \\ \hline
\end{tabular}
\caption{We examined to what extent the individual
embeddings store morphological information. To quantify this,
we treated the problem as supervised multi-way classification
with the embedding as the input and the morphological tag 
as the output to predict. Note that ``All Types'' refers 
to all types in the training corpus and ``No Tags'' refers
to the subset of types, whose morphological tag was {\em not} seen 
by Morph-LBL at training time.  }
\label{table:knn-results}
\end{table}

We first investigated whether the embeddings learned by Morph-LBL do
indeed encode morphological information. For each word, we selected
the most frequently occurring morphological tag for that word (ties
were broken randomly). We then treated the problem of labeling a
word-embedding with its most frequent morphological tag as a multi-way
classification problem. We trained a $k$ nearest neighbors classifier where $k$
was optimized on development data. We used the  
\texttt{scikit-learn} library \cite{pedregosa2011scikit} on all types in
the vocabulary with 10-fold cross-validation, holding out 10\% of
the data for testing at each fold and an additional 10\% of training as a development
set. The results displayed in table
\ref{table:knn-results} are broken down by whether MorphLBL observed
the morphological tag at training time or not. We see that embeddings from Morph-LBL do
store the proper morphological analysis at a much higher rate
than both the vanilla LBL and \textsc{word2vec}. 

Word-embeddings,
however, are often trained on massive amounts of unlabeled data. To this
end, we also explored on how \textsc{word2vec} itself encodes morphology,
when trained on an order of magnitude more data. Using the same experimental setup
as above, we trained \textsc{word2vec} on the {\em union} of the TIGER German corpus
and German section of Europarl \cite{koehn2005europarl} for a total of $\approx$ 45 million tokens. 
Looking only at those types found in TIGER, we found that the $k$-NN classifier predicted
the correct tag with $\approx$ 22\% accuracy 
(not shown in the table).

\subsection{Experiment 2: \textsc{MorphoDist}}

\begin{figure}[!t]
\centering
\includegraphics[width=0.5\textwidth]{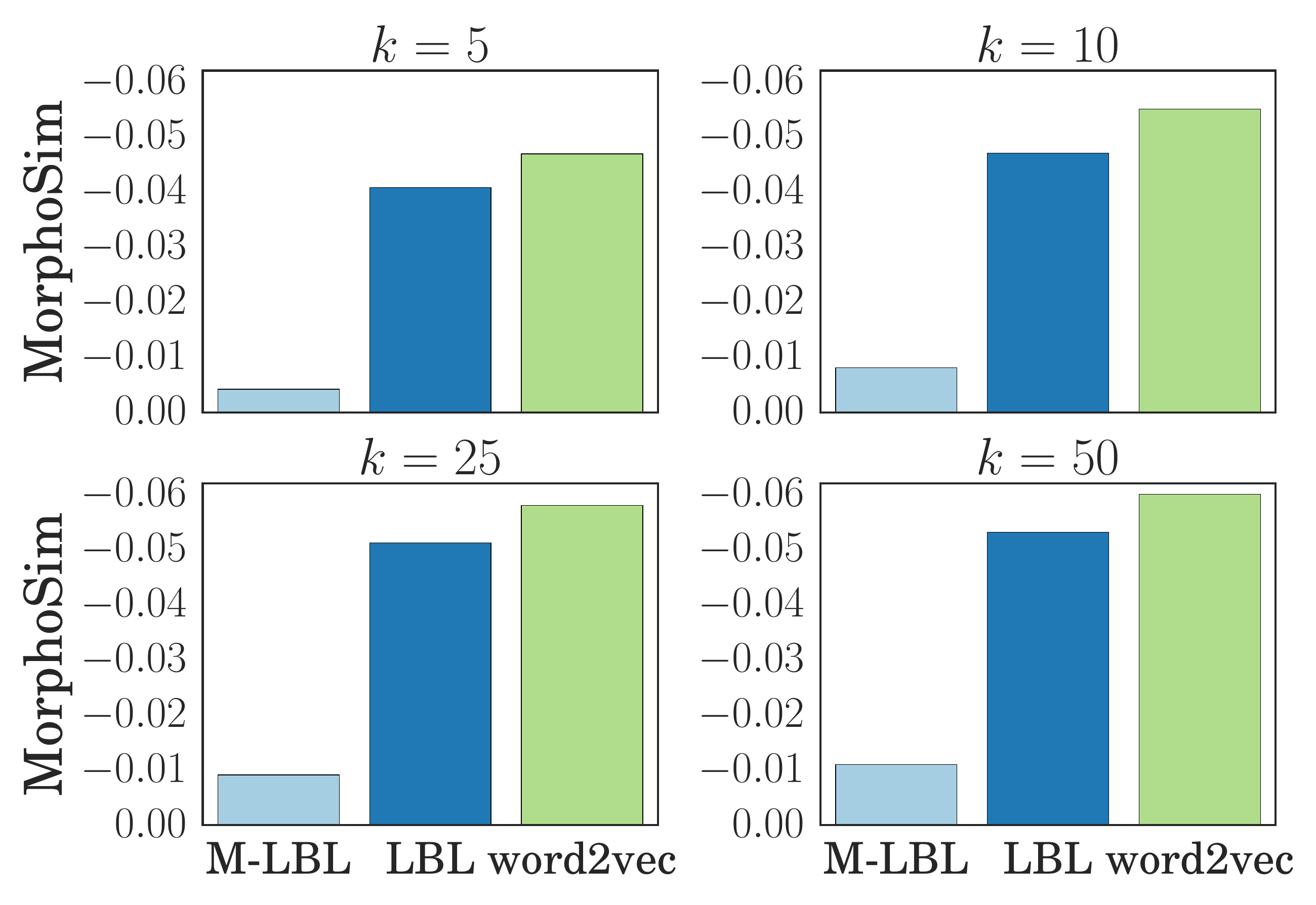}
\caption{Results for the
\morphosim measure for $k \in \{5,10,25,50\}$.  Lower
\morphosim values are better---they indicate that the nearest
neighbors of each word are closer morphologically.}
\label{fig:morph-sim}
\end{figure}

We also evaluated the three types of embeddings using the
\morphosim metric introduced in section
\ref{sec:morpho-sim}. This metric roughly tells us how similar each
word is to its neighbors, where distance is measured in the Hamming
distance between morphological tags. We only evaluated on words that
MorphLBL did {\em not} observe at training time to get a fair idea of
how well our model has managed to encode morphology purely from the
contextual signature. Figure \ref{fig:morph-sim} reports results for
$k \in \{5,10,25,50\}$ nearest neighbors. We see that the values of
$k$ studied do not affect the metric---the closest 5 words are about
as similar as the closest 50 words.  We see again that the Morph-LBL
embeddings generally encode morphology better than the baselines.

\subsection{Discussion} 
The superior performance of Morph-LBL over both the original LBL 
and \textsc{word2vec} under both evaluation metrics is not surprising as we provide
our model with annotated data at training time. That the LBL
outperforms \textsc{word2vec} is also not surprising. The LBL looks at a
local history thus making it more amenable to learning
syntactically-aware embeddings than \textsc{word2vec}, whose
skip-grams often look at non-local context.  

What is of
interest, however, is Morph-LBL's ability to robustly maintain
morphological relationships only making use of the distributional
signature, {\em without} word-internal features. \emph{This result
shows that in large corpora, a large portion of morphology can be 
extracted through contextual similarity.} 

\enote{hs}{i tried to emphasize this typographically since
  i'm afraid this very important pint may get lost

in hindsight it would have been better to switch the order
of the two paragraphs: first say what's intresting, then say
whats' not interstign

but better not to make changes this late}

\section{Conclusion and Future Work}
We described a new model, Morph-LBL, for the semi-supervised induction
of morphologically guided embeddings. The combination of
morphologically annotated data with raw text allows us to train
embeddings that preserve morphological relationships among words.  Our
model handily outperformed two baselines trained on the same corpus.

While contextual signatures provide a strong cue for morphological
proximity, orthographic features are also requisite for a strong
model. Consider the words \word{loving} and
\word{eating}. Both are likely to occur after \word{is}/\word{are} and thus their local contextual signatures are likely to be
similar.  However, perhaps an equally strong signal is that the two
words end in the same substring \word{ing}. Future work will handle
such integration of  character-level features.

We are interested in the application of our embeddings
to morphological tagging and other tasks.  Word-embeddings have proven themselves as
useful features in a variety of tasks in the NLP
pipeline. Morphologically-driven embeddings have the potential to
leverage raw text in a way state-of-the-art morphological taggers
cannot, improving tagging performance downstream.

\section*{Acknowledgements}
This material is based upon work supported by a Fulbright fellowship
awarded to the first author by the German-American Fulbright
Commission and the National Science Foundation under Grant
No. 1423276. 
The second author was supported by Deutsche
Forschungsgemeinschaft (grant DFG
SCHU 2246/10-1).
We thank Thomas M{\"u}ller for several insightful
discussions on morphological tagging and Jason Eisner
for discussions about experimental design. Finally, we thank the
anonymous reviewers for their many helpful comments.

\bibliographystyle{naaclhlt2015}
\bibliography{morphological_lbl}

\end{document}